\documentclass[journal]{IEEEtran}
\usepackage{amsmath,epsfig}
\graphicspath{{./}{figureHDQAR/}}
\usepackage{multirow}
\usepackage{graphicx,epsfig, epstopdf,subfigure}
\usepackage{algorithm, algorithmic,url}
  \usepackage{mathrsfs,amssymb,amsmath,cite, color,bm,tabu,comment}
  \usepackage[table]{xcolor}

\begin{document}
%
\title{Color Face Recognition using High-Dimension Quaternion-based Adaptive Representation}
%
%
%

\author{Qingxiang Feng, 
        Yicong~Zhou,~\IEEEmembership{Senior~Member,~IEEE,}
\thanks{This work was supported in part by the Macau Science and Technology Development Fund under Grant FDCT/016/2015/A1 and by the Research Committee at University of Macau under Grants MYRG2014-00003-FST and MYRG2016-00123-FST.
(Corresponding author is Yicong Zhou.)}
\thanks{All authors are with the Department of Computer and Information Science, University
of Macau, Macau 999078, China
(e-mail: fengqx1988@gmail.com; yicongzhou@umac.mo).}
}

\maketitle

\begin{abstract}

Recently, quaternion collaborative representation-based classification (QCRC) and quaternion sparse representation-based classification (QSRC) have been proposed for color face recognition. They can obtain correlation information among different color channels. However, their performance is unstable in different conditions. For example, QSRC performs better than than QCRC on some situations but worse on other situations. To benefit from quaternion-based $e_2$-norm minimization in QCRC and quaternion-based $e_1$-norm minimization in QSRC, we propose the quaternion-based adaptive representation (QAR) that uses a quaternion-based $e_p$-norm minimization ($1 \le p \le 2$) for color face recognition. To obtain the high dimension correlation information among different color channels, we further propose the high-dimension quaternion-based adaptive representation (HD-QAR). The experimental results demonstrate that the proposed QAR and HD-QAR achieve better recognition rates than QCRC, QSRC and several state-of-the-art methods.

\end{abstract}


\IEEEpeerreviewmaketitle

\section{Introduction}

\IEEEPARstart{F}{ace} recognition systems are relied on classification methods. Recently, representation-based classification methods \cite{SRC,SSPC,AOSE,CWKLR,NFLS,DTSL-LRSR,TC-NFP} have attracted more attention and obtained good performance in face recognition. Two typical representation-based classification methods are sparse representation-based classification (SRC)\cite{SRC}\cite{SRC1} and collaborative representation based classification (CRC) \cite{CRC}. For color face recognition, SRC and CRC independently use color channels of color face images. As a result, they fails to consider the structural correlation information among different color channels. The previous research works \cite{Quaternion01,Quaternion02,Quaternion03,Quaternion04,Quaternion05,Quaternion06} proved that quaternion can obtain the information among different channels of the color images by representing three color channels as three imaginary parts. For example, \cite{Quaternion05} proposed a quaternion-based structural similarity index and used it to assess the quality of color images. \cite{VSR-QMA} proposed a vector sparse representation using quaternion matrix analysis and successfully applied it to various color image processing. \cite{QCRC-QSRC} proposed two methods to obtain correlation information among red, green and blue channels for color face recognition. These two methods are named quaternion collaborative representation-based classification (QCRC) and quaternion sparse representation-based classification (QSRC). QCRC obtains the sparse coefficient by solving quaternion-based $e_2$-norm minimization problem while QSRC solves quaternion $e_1$-norm minimization problem to compute the sparse coefficient. From the results in \cite{QCRC-QSRC}, we find an interesting phenomenon that QSRC obtains better performance than QCRC on some situations while QCRC performs better than QSRC on other situations. Table \ref{t_motivationQAR} proves this. That is, quaternion-based $e_2$-norm minimization and quaternion-based $e_1$-norm minimization both have their advantages.

Motivated by this, we want to design a quaternion-based $e_p$-norm minimization ($1 \le p \le 2$) to solve the sparse optimization problem such that it can benefit both from quaternion-based $e_2$-norm minimization in QCRC and quaternion-based $e_1$-norm minimization in QSRC. Moreover, \cite{KSR} shows that high dimension information in kernel space is helpful for sparse. We also want to design a kernel-quaternion-based transformation that can obtain the high dimension correlation information among different color channels. Our main contributions are described as follows.
\begin{itemize}
  \item To benefit both from quaternion-based $e_2$-norm minimization in QCRC and quaternion-based $e_1$-norm minimization in QSRC, we propose quaternion-based adaptive representation (QAR) method. QAR uses a quaternion-based $e_p$-norm minimization ($1 \le p \le 2$) that can adaptively combine quaternion-based $e_2$-norm minimization and quaternion-based $e_1$-norm minimization.
  \item To our best knowledge, the high-dimension quaternion has not been applied. Thus, we propose the high-dimension quaternion-based adaptive representation (HD-QAR). It is able to obtain high-dimension correlation information among different color channels.
  \item Because the product of high-dimension quaternion is non-communicative, it is quite difficult to solve the high-dimension quaternion-based optimization problem in HD-QAR.  Thus, we propose the quaternion-based kernel matrix and two high-dimension-quaternion operators to solve optimization problem in HD-QAR be easy.
  \item Extensive experiments on six well-known color face databases show that the proposed methods have better performance than several state-of-the-art methods.
\end{itemize}

\begin{table} [t]
\caption{Notation Summary} \label{t_NotationSummary}
\begin{center}
\begin{tabular}{|l|l|}
\hline
Notation & Description \\
\hline\hline
$X$  & Entire training set  \\
$X_c$  & All samples of the $c^{\mbox {th}}$ class  \\
$x_i$  & The $i^{\mbox{th}}$ sample of $X$   \\
$q$  & Dimension of a sample  \\
$N_c$  & Number of samples of the $c^{\mbox{th}}$ class \\
$M$  & Number of classes  \\
$L$  & Number of samples of $X$  \\
$x$  & A testing sample \\
$\dot X$  & Matrix-based quaternion  \\
$\dot x$  & Vector-based quaternion \\
\hline
\end{tabular}
\end{center}
\end{table}

\section{Related Work} \label{sect_II}

This section reviews the related works including quaternion, kernel and representation-based methods.

\subsection{Quaternion Algebra}

Hamilton proposed the quaternion space that has three imagery units $i$, $j$ and $k$ \cite{QuaternionHamilton}. These three imagery units are defined as
\begin{equation}
{i^2} = {j^2} = {k^2} = ijk =  - 1
\end{equation}
A quaternion $\dot q$ is described by
\begin{equation}
\dot q = {q_0} + {q_1}i + {q_2}j + {q_3}k
\end{equation}
where $\begin{array}{*{20}{c}}
{{q_0},}&{{q_1},}&{{q_2},}&{{q_3}}
\end{array}$ are real numbers.
The conjugate quaternion of $\dot q$ is
\begin{equation}
\bar \dot q = {q_0} - {q_1}i - {q_2}j - {q_3}k
\end{equation}
The modulus of quaternion $\dot q$ is
\begin{equation}
|\dot q| = \sqrt {\dot q\bar \dot q}  = \sqrt {q_0^2 + q_1^2 + q_2^2 + q_3^2}
\end{equation}
The addition operation of two quaternions $\dot p$ and $\dot q$ is
\begin{equation}
\dot q + \dot p = ({q_0} + {p_0}) + ({q_1} + {p_1})i + ({q_2} + {p_2})j + ({q_3} + {p_3})k
\end{equation}
The multiplication operation of two quaternions $\dot p$ and $\dot q$ is
\begin{equation}
\begin{array}{l}
\dot q\dot p = ({q_0}{p_0} - {q_1}{p_1} - {q_2}{p_2} - {q_3}{p_3})\\
\;\;\;\;\;\;\; + ({q_0}{p_1} + {q_1}{p_0} + {q_2}{p_3} - {q_3}{p_2})i\\
\;\;\;\;\;\;\; + ({q_0}{p_2} - {q_1}{p_3} + {q_2}{p_0} + {q_3}{p_1})j\\
\;\;\;\;\;\;\; + ({q_0}{p_3} + {q_1}{p_2} - {q_2}{p_3} + {q_3}{p_0})k
\end{array}
\end{equation}

\subsection{Kernel trick}

The kernel techniques can map the data to a high-dimension space. Using the kernel trick, we can obtain the high dimension information without computing the mapping explicitly. In this paper, we use a well-known kernel called the Gaussian radial basis function (RBF) kernel. It can be represented as
\begin{equation}
k(a,b) = \phi {(a)^T}\phi (b) = \exp (\frac{{ - ||a - b||_{}^2}}{\delta })
\end{equation}
where $a$ and $b$ are any two original samples, $\delta $ is a parameter. In the kernel trick, $\phi (*)$ is unknown. We can access the feature space only via $k(*,*)$ .

\subsection{Representation-based Methods}

\emph{Wright et al.} proposed sparse representation classification (SRC) in 2009. It represents the testing sample by the linear combination of training samples of all classes. SRC solves $e_1$-minimization optimization problem as
\begin{equation}
\min \;||\alpha |{|_1}\;\;\;\;s.t.\;\;\;x = X\alpha
\end{equation}
To reduce the computation cost and obtain the better representation, \emph{Zhang et al.} proposed the collaborative representation based classification (CRC). In CRC, the authors argued that the collaborative representation should be better than the linear combination of training samples. CRC solves the $e_2$-minimization optimization problem as
\begin{equation}
\min \;||\alpha ||_2^2\;\;\;\;s.t.\;\;\;x = X\alpha
\end{equation}
They are two typical sparse-based methods. A lot modifications of SRC/CRC have been proposed for various visual recognition tasks \cite{TPSR,ASRC,Tracelasso,KRDU,KCSR}. For example, \emph{Gao et al.} denoted that the nonlinear high dimensional features are useful for sparse representation, and then proposed kernel-based sparse representation classification (KSRC) \cite{KSR,KSR1},
\begin{equation}
\min \;||\alpha |{|_1}\;\;\;\;s.t.\;\;\;\Phi (x) = \Phi (X)\alpha
\end{equation}
\emph{Wang et al.} proposed kernel-based collaborative representation classification (KCRC) \cite{KCRC}, which is described as
\begin{equation}
\min \;||\alpha |{|_2}\;\;\;\;s.t.\;\;\;\Phi (x) = \Phi (X)\alpha
\end{equation}
\emph{Zou et al.} denoted that quaternion can help sparse representation to obtain the correlation information among different color channels. Then, they proposed quaternion sparse representation classification (QSRC) and quaternion collaborative representation based classification (QCRC). QSRC is described as
\begin{equation}
\min \;||\dot \alpha |{|_1}\;\;\;\;s.t.\;\;\;\dot x = \dot X\dot \alpha
\end{equation}
QCRC is described as
\begin{equation}
\min \;||\dot \alpha |{|_2}\;\;\;\;s.t.\;\;\;\dot x = \dot X\dot \alpha
\end{equation}

\begin{figure} [t]
\begin{center}
\subfigure{\includegraphics[width=3.0in]{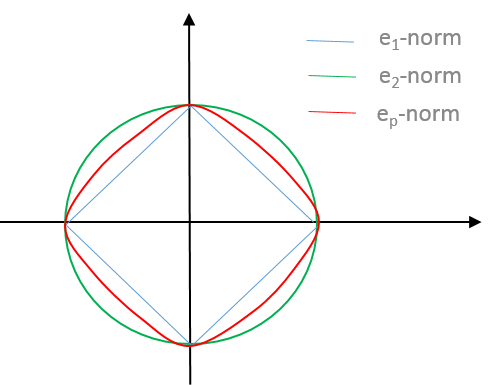}}
\end{center}
\vspace{-0.05in} \caption{ Motivation: Because quaternion-based $e_2$-norm and $e_1$-norm both have their own advantages, we want to design a quaternion-based $e_p$-norm ($1 \le p \le 2$) that can adaptively combine quaternion-based $e_2$-norm and $e_1$-norm and benefit from both. The above is the geometric interpretation of $e_p$-norm, $e_2$-norm and $e_1$-norm. } \label{f_motivationQAR}
\end{figure}

\begin{table} [t]
\caption{Mean results of Table III in \cite{QCRC-QSRC}} \label{t_motivationQAR}
\begin{center}
\begin{tabular}{|l|c|c|}
\hline
Method & QCRC & QSRC \\
\hline\hline
1  & 48.70  & 51.57  \\
2  & 67.94 & 68.67  \\
3  & 75.80 & 75.41   \\
4  & 78.99 & 78.41  \\
\hline
\end{tabular}
\end{center}
\end{table}

\section{Proposed QAR}
In this section, we propose a new method, called quaternion-based adaptive representation (QAR). We first introduce its motivation. Then, we define and analyze its objective function. Afterwards, the solution of this objective function is given. Last, we describe the classification rule and summary this method.

\subsection{Motivation}

  To consider the structural correlation information among different color channels, \cite{QCRC-QSRC} proposed two methods for color face recognition. They are quaternion collaborative representation-based classification (QCRC) using $e_1$-norm minimization and quaternion sparse representation-based classification (QSRC) using $e_2$-norm minimization. Table \ref{t_motivationQAR} lists some results copied from \cite{QCRC-QSRC}. From Table \ref{t_motivationQAR} and \cite{QCRC-QSRC}, we find that QSRC obtains the better performance than QCRC on some situations while QCRC obtains the better performance than QSRC on other situations.  This means that quaternion-based $e_2$-norm minimization and quaternion-based $e_1$-norm minimization both have their own advantages. Motivated by this, this paper designs a quaternion-based $e_p$-norm minimization ($1 \le p \le 2$) to solve the sparse optimization problem such that it can benefit from quaternion-based $e_2$-norm minimization in QCRC and quaternion-based $e_1$-norm minimization in QSRC. We give an example in Fig. \ref{f_motivationQAR} to describe the geometric interpretation of $e_p$-norm, $e_2$-norm and $e_1$-norm.

\subsection{Objective function of QAR}
This paper tries to propose a quaternion-based $e_p$-norm that can adaptively combine quaternion-based $e_2$-norm and $e_1$-norm and benefit from both. Our objective function can be described as follows.
\begin{equation} \label{eq_QARob1}
\min \;||\dot \alpha |{|_p}\;\;\;\;s.t.\;\;\;\dot x = \dot X\dot \alpha
\end{equation}
where $1 \le p \le 2$.
Now, we face a problem how to obtain a better $e_p$-norm. To solve this, we give a new objective function that can be treated as $e_p$-norm as
\begin{equation} \label{eq_QARob2}
\begin{array}{l}
\min ||\dot XDiag(\dot \alpha )|{|_*}\\
\begin{array}{*{20}{c}}
{s.t.}&{\dot x = }
\end{array}\dot X\dot \alpha
\end{array}
\end{equation}
where $|| \bullet |{|_*}$ means the nuclear or trace norm. \\
Because the product of quaternions are non-communicative, it is quite difficult to solve the quaternion-based optimization problem in QAR. Following \cite{QCRC-QSRC}, we give two quaternion-based operations $\Re (*)$ and $\aleph (*)$ . They are defined as follows. Given a matrix-based quaternion  $\dot V = {V_0} + {V_1}i + {V_2}j + {V_3}k$ . $\Re (\dot V)$  is equal to
\begin{equation}
{\Re ^{}}(\dot V): = \left[ {\begin{array}{*{20}{c}}
{{V_0}}&{ - {V_1}}&{ - {V_2}}&{ - {V_3}}\\
{{V_1}}&{{V_0}}&{ - {V_3}}&{{V_2}}\\
{{V_2}}&{{V_3}}&{{V_0}}&{ - {V_1}}\\
{{V_3}}&{ - {V_2}}&{{V_1}}&{{V_0}}
\end{array}} \right]
\end{equation}
Given a vector-based quaternion $\dot v = {v_0} + {v_1}i + {v_2}j + {v_3}k$.  $\aleph (\dot v)$ is equal to
\begin{equation}
\aleph \left( {\dot v} \right) = {\left[ {\begin{array}{*{20}{c}}
{v_0^T}&{v_1^T}&{v_2^T}&{v_3^T}
\end{array}} \right]^T}
\end{equation}
Using the two quaternion-based operations $\Re (*)$ and $\aleph (*)$, the objective function in Eq. (\ref{eq_QARob1}) can be rewritten as
\begin{equation} \label{eq_QARob3}
\min \;||\aleph (\dot \alpha )|{|_p}\;\;\;\;s.t.\;\;\;\aleph (\dot x) = \Re (\dot X)\aleph (\dot \alpha )
\end{equation}
and the objective function in Eq. (\ref{eq_QARob2}) can be rewritten as
\begin{equation} \label{eq_QARob4}
\begin{array}{l}
\min ||\Re (\dot X)Diag(\aleph (\dot \alpha ))|{|_*}\\
\begin{array}{*{20}{c}}
{s.t.}&{\aleph (\dot x) = }
\end{array}{\Re ^{}}(\dot X)\aleph (\dot \alpha )
\end{array}
\end{equation}
Next, we will explain why Eq. (\ref{eq_QARob4}) can be treated as $e_p$-norm ($1 \le p \le 2$).

\subsection{Analysis of objective function}

 This section analyzes the objective function and explain why $\min ||\Re (\dot X)Diag(\aleph (\dot \alpha ))|{|_*}$ in Eq. (\ref{eq_QARob4}) is equivalent to $\min \;||\aleph (\dot \alpha )|{|_p}\;$ in Eq (\ref{eq_QARob3}). In order to explain the proposed objective function can be treated as $e_p$-norm minimization ($1 \le p \le 2$), we give two examples. Suppose that the subjects (columns) of $\Re (\dot X)$  are orthogonal and different from each other, that is, $\Re {(\dot X)^T}\Re (\dot X) = I$ , where $I$ is an identity matrix. We define $\Omega  = ||\Re (\dot X)Diag(\aleph (\dot \alpha ))|{|_*}$ and its decomposition is
\begin{equation} \label{eq_QARaof1}
\begin{array}{l}
\Omega  = Tr{\left[ {{{\left( {\Re (\dot X)Diag(\aleph (\dot \alpha ))} \right)}^T}\left( {\Re (\dot X)Diag(\aleph (\dot \alpha ))} \right)} \right]^{{\raise0.7ex\hbox{$1$} \!\mathord{\left/
 {\vphantom {1 2}}\right.\kern-\nulldelimiterspace}
\!\lower0.7ex\hbox{$2$}}}}\\
 ~~= Tr{\left[ {{{\left( {Diag(\aleph (\dot \alpha ))} \right)}^T}\left( {Diag(\aleph (\dot \alpha ))} \right)} \right]^{{\raise0.7ex\hbox{$1$} \!\mathord{\left/
 {\vphantom {1 2}}\right.\kern-\nulldelimiterspace}
\!\lower0.7ex\hbox{$2$}}}}\\
 ~~= ||\aleph (\dot \alpha )|{|_1}
\end{array}
\end{equation}
Based on Eq. (\ref{eq_QARaof1}), we know that the objective function in Eq. (\ref{eq_QARob4}) is equivalent to $e_1$-norm minimization problem. \\
On the contrary, we suppose that the subjects (columns) of $\Re (\dot X)$ are same to the first column of $\Re (\dot X)$ , that is $\Re {(\dot X)^T}\Re (\dot X) = {11^T}$ ( $1$ is a vector that all elements are one). We define the first column of $\Re (\dot X)$ as ${\Re (\dot X)}_1$  and  The decomposition is
\begin{equation} \label{eq_QARaof2}
\Omega  = ||\Re {(\dot X)_1}{1^T}Diag(\aleph (\dot \alpha ))|{|_*}
\end{equation}
Suppose that the rank of a matrix $A \in {R^{m \times n}}$  is $r$. We can obtain the inequalities \cite{golub2012matrix}\cite{horn2012matrix} as follows. \footnote{\url{https://en.wikipedia.org/wiki/Matrix_norm}}
\begin{equation} \label{eq_QARaof3}
||A|{|_F} \le ||A|{|_*} \le \sqrt r ||A|{|_F}
\end{equation}
where $|| \bullet |{|_F}$ is the Frobenius norm. Because the rank of $\Re {(\dot X)_1}{1^T}Diag(\aleph (\dot \alpha ))$ is one ($r=1$), $\Omega  = ||\Re (\dot X)Diag(\aleph (\dot \alpha ))|{|_*}$ can be re-wrote as
\begin{equation} \label{eq_QARaof4}
\begin{array}{c}
\Omega  = ||\Re {(\dot X)_1}{1^T}Diag(\aleph (\dot \alpha ))|{|_*} ~~~~~~~~~~~\\
=||\Re {(\dot X)_1}{1^T}Diag(\aleph (\dot \alpha ))|{|_F}  ~~~~~~~~ \\
 = ||\Re {(\dot X)_1}|{|_2}||{1^T}|{|_2}||Diag(\aleph (\dot \alpha ))|{|_2}\\
  =||\aleph (\dot \alpha )|{|_2}  ~~~~~~~~~~~~~~~~~~~~~~~~~~~~~
\end{array}
\end{equation}
In the real applications, the subjects (columns) of data set $\Re (\dot X)$  are not different each other or same to each other. Consider the Eqs. (\ref{eq_QARaof1}) and (\ref{eq_QARaof4}), our objective function could be treated as a combination of the quaternion-based $e_1$-norm minimization in QSRC and quaternion-based $e_2$-norm minimization in QCRC. That is, Eq. (\ref{eq_QARob4}) can be treated as quaternion-based $e_p$-norm. Our objective function could benefit from both quaternion-based $e_1$-norm minimization and quaternion-based $e_2$-norm minimization based on the correction of the subjects (columns) in the data set $\Re (\dot X)$ .

\begin{algorithm*}[htbp]
  \begin{algorithmic}[1]
  \REQUIRE The original training set $X$ with $L$ samples and a testing sample  $x$. Initialize the value of $Z,z,\alpha ,{m_1},{M_2},u,\rho ,\varepsilon $ and $u_{max}$
  \ENSURE  Label of the testing sample $x$.
  \vspace{3pt}
  \STATE  Constitute the objective function using the original training set $X$ and testing sample $x$ as
  \[\aleph {(\dot \alpha )^*} = \arg \min ||\aleph (\dot x) - \Re (\dot X)\aleph (\dot \alpha )|{|_1} + \lambda ||\Re (\dot X)Diag(\aleph (\dot \alpha ))|{|_*}\]
  \STATE Transform the objective function to a Lagrange multiplier problem as
  \[\begin{array}{l}
L(Z,z,\aleph (\dot \alpha )) = \lambda ||Z|{|_*} + ||z|{|_1} + m_1^T(a - \Re (\dot X)\aleph (\dot \alpha ) - z) + Tr[M_2^T(Z - \Re (\dot X)Diag(\aleph (\dot \alpha )))]\\
{\kern 1pt} {\kern 1pt} {\kern 1pt} {\kern 1pt} {\kern 1pt} {\kern 1pt} {\kern 1pt} {\kern 1pt} {\kern 1pt} {\kern 1pt} {\kern 1pt} {\kern 1pt} {\kern 1pt} {\kern 1pt} {\kern 1pt} {\kern 1pt} {\kern 1pt} {\kern 1pt} {\kern 1pt} {\kern 1pt} {\kern 1pt} {\kern 1pt} {\kern 1pt} {\kern 1pt} {\kern 1pt} {\kern 1pt} {\kern 1pt} {\kern 1pt} {\kern 1pt} {\kern 1pt} {\kern 1pt} {\kern 1pt} {\kern 1pt} {\kern 1pt} {\kern 1pt} {\kern 1pt} {\kern 1pt} {\kern 1pt} {\kern 1pt} {\kern 1pt} {\kern 1pt} {\kern 1pt} {\kern 1pt} {\kern 1pt} {\kern 1pt} {\kern 1pt} {\kern 1pt} {\kern 1pt} {\kern 1pt} {\kern 1pt} {\kern 1pt} {\kern 1pt} {\kern 1pt} {\kern 1pt} {\kern 1pt} {\kern 1pt} {\kern 1pt} {\kern 1pt} {\kern 1pt} {\kern 1pt}  + \frac{u}{2}(||\aleph (\dot x) - \Re (\dot X)\aleph (\dot \alpha ) - z||_2^2 + ||Z - \Re (\dot X)Diag(\aleph (\dot \alpha ))||_F^2)
\end{array}\]
  \WHILE {until convergence}
  \vspace{3pt}
  \STATE  Update $Z$ by fixing the others as $Z* = \arg \mathop {\min }\limits_Z \frac{\lambda }{u}||Z|{|_*} + \frac{1}{2}||Z - (\Re (\dot X)Diag(\aleph (\dot \alpha )) - \frac{1}{u}M_2^{})||_F^2$
  \vspace{3pt}
  \STATE  Update $\aleph (\dot \alpha )$  by fixing the others as \[\begin{array}{c}
\aleph (\dot \alpha )* = {(\Re {(\dot X)^T}\Re (\dot X) + Diag(diag(\Re {(\dot X)^T}\Re (\dot X))))^{ - 1}}\Re {(\dot X)^T}(\frac{1}{u}{m_1} + \Re (\dot X)(\aleph (\dot x) - z))\\
 + {(\Re {(\dot X)^T}\Re (\dot X) + Diag(diag(\Re {(\dot X)^T}\Re (\dot X))))^{ - 1}}diag(\Re {(\dot X)^T}(\frac{1}{u}{M_2} + Z))
\end{array}\]
  \STATE  Update $z$ by fixing the others as $z* = \arg \mathop {\min }\limits_z \frac{1}{u}||z|{|_1} + \frac{1}{2}||z - (\aleph (\dot x) - \Re (\dot X)\aleph (\dot \alpha ) + \frac{1}{u}{m_1})||_2^{}$
  \vspace{3pt}
  \STATE  The multipliers $m_1,M_2$ are updated by ${m_1} = {m_1} + u(\aleph (\dot x) - \Re (\dot X)\aleph (\dot \alpha ) - z)$ and ${M_2} = {M_2} + u(Z - \Re (\dot X)Diag(\aleph (\dot \alpha )))$
  \vspace{3pt}
  \STATE   Update the parameter $u$ by $u = \min (\rho u,{u_{\max }})$
  \vspace{3pt}
  \STATE   Check convergence by $||(\aleph (\dot x) - \Re (\dot X)\aleph (\dot \alpha ) - z|{|_\infty } \le \varepsilon $ and $||E - ADiag(\beta )|{|_\infty } \le \varepsilon $
  \vspace{3pt}
  \ENDWHILE
  \vspace{3pt}
  \STATE   Compute the distance ${d_c} = ||\aleph (\dot x) - \Re ({\dot X_c})\aleph ({\dot \alpha _c})||$ and classify the testing sample $x$  by $c* = \arg \min ({d_c})$

\end{algorithmic}
\caption{Proposed Quaternion-based Adaptive Representation (QAR) Method}\label{algorithm1}
\end{algorithm*}

\subsection{Solution of objective function}

Based on the optimization methods in \cite{candes2011robust} and \cite{liu2010robust}, we use the inexact augmented Lagrange multipliers (IALM) [40] method to solve the proposed $e_p$-norm minimization. We convert the minimization problem in Eq.(\ref{eq_QARob4}) to the following optimization problem as
\begin{equation} \label{eq_QARob5}
\begin{array}{l}
\mathop {\min }\limits_{Z,z,\aleph (\dot \alpha )} ||Z|{|_*} + ||z|{|_1}\\
\begin{array}{*{20}{c}}
{s.t.}&{z = }
\end{array}\aleph (\dot x) - {\Re ^{}}(\dot X)\aleph (\dot \alpha )\\
\begin{array}{*{20}{c}}
{{\kern 1pt} {\kern 1pt} {\kern 1pt} {\kern 1pt} {\kern 1pt} {\kern 1pt} {\kern 1pt} {\kern 1pt} {\kern 1pt} {\kern 1pt} }&{Z = \Re (\dot X)Diag(\aleph (\dot \alpha ))}
\end{array}
\end{array}
\end{equation}
We further convert the optimization problem in Eq. (\ref{eq_QARob5}) to a augmented Lagrange multiplier problem as follows.
\begin{equation} \label{eq_QARsof1}
\begin{array}{l}
L(Z,z,\aleph (\dot \alpha )) = \lambda ||Z|{|_*} + m_1^T(\aleph (\dot x) - \Re (\dot X)\aleph (\dot \alpha ) - z)\\
{\kern 1pt}  + ||z|{|_1} + Tr[M_2^T(Z - \Re (\dot X)Diag(\aleph (\dot \alpha )))]\\
{\kern 1pt}  + \frac{u}{2}(||a - \Re (\dot X)\aleph (\dot \alpha ) - z||_2^2 + ||Z - \Re (\dot X)Diag(\aleph (\dot \alpha ))||_F^2)
\end{array}
\end{equation}
where $u > 0$ is a parameter, $m_1$  and $M_2$ are Lagrange multipliers.
We can alternatively optimize the variates $Z,z$ and $\aleph (\dot \alpha )$  by fixing others. The detailed optimization processes of $Z,z$ and $\aleph (\dot \alpha )$ are described as follows.\\
 Update $Z$ by fixing $z$ and $\aleph (\dot \alpha )$ . It is to solve the following minimization problem as
 \begin{equation} \label{eq_QARsof2}
\begin{array}{l}
Z* = \arg \mathop {\min }\limits_Z L(Z,z,\aleph (\dot \alpha ))\\
 = \arg \mathop {\min }\limits_Z \lambda ||Z|{|_*} + Tr(M_2^TZ) \\
~~~~~~~~~~~~ + \frac{u}{2}||Z - \Re (\dot X)Diag(\aleph (\dot \alpha ))||_F^2\\
 = \arg \mathop {\min }\limits_Z \frac{\lambda }{u}||Z|{|_*} + \frac{1}{2}||Z - (\Re (\dot X)Diag(\aleph (\dot \alpha )) - \frac{1}{u}M_2^{})||_F^2
\end{array}
\end{equation}
 We can approximately solve the above minimization problem in Eq. (\ref{eq_QARsof2}) using the singular value thresholding (SVT) operator \cite{SVToperator}.\\
Then, update $\aleph (\dot \alpha )$  by fixing $z$ and $Z$. It is to solve the following minimization problem as.
 \begin{equation} \label{eq_QARsof3}
\begin{array}{l}
\aleph (\dot \alpha )* = \arg \mathop {\min }\limits_{\aleph (\dot \alpha )} L(Z,z,\aleph (\dot \alpha ))\\
 = \arg \mathop {\min }\limits_{\aleph (\dot \alpha )}  - m_1^TJ\aleph (\dot \alpha ) - Tr(M_2^TJDiag(\aleph (\dot \alpha )))\\
 ~~~~+ \frac{u}{2}(\aleph {(\dot \alpha )^T}{J^T}J\aleph (\dot \alpha ) - 2{(\aleph (\dot x) - z)^T}J\aleph (\dot \alpha ) \\
 ~~~~- 2Tr({Z^T}JDiag(\aleph (\dot \alpha )))\\
 ~~~~+ Tr({(JDiag(\aleph (\dot \alpha )))^T}JDiag(\aleph (\dot \alpha ))))\\
 = \arg \mathop {\min }\limits_{\aleph (\dot \alpha )} \frac{u}{2}\aleph {(\dot \alpha )^T}({J^T}J + Diag(diag({J^T}J)))\aleph (\dot \alpha ) - \\
{({J^T}{m_1} + u{J^T}J(\aleph (\dot x) - z) + diag(M_2^TJ + u{Z^T}J))^T}\aleph (\dot \alpha )
\end{array}
\end{equation}
 where $J=\Re (\dot X)$. The above minimization problem in Eq. (\ref{eq_QARsof3}) can be solved by
\begin{equation} \label{eq_QARsof4}
\begin{array}{l}
\aleph (\dot \alpha )* =\\
 {({J^T}J + Diag(diag({J^T}J)))^{ - 1}}{J^T}(\frac{1}{u}{m_1} + J(\aleph (\dot \alpha ) - z))\\
 + {({J^T}J + Diag(diag({J^T}J)))^{ - 1}}diag({J^T}(\frac{1}{u}{M_2} + Z))
\end{array}
\end{equation}
Next, update $z$ by fixing $Z$ and  $\aleph (\dot \alpha )$. It is to solve the following minimization problem as
\begin{equation} \label{eq_QARsof5}
\begin{array}{l}
z* = \arg \mathop {\min }\limits_z L(Z,z,\aleph (\dot \alpha ))\\
 = \arg \mathop {\min }\limits_z ||z|{|_1} - m_1^Tz + \frac{u}{2}||a - \Re (\dot X)\aleph (\dot \alpha ) - z||_F^2\\
 = \arg \mathop {\min }\limits_z \frac{1}{u}||z|{|_1} + \frac{1}{2}||z - (a - \Re (\dot X)\aleph (\dot \alpha ) + \frac{1}{u}{m_1})||_2^2
\end{array}
\end{equation}
We can solve the minimization problem in Eq. (\ref{eq_QARsof5}) by using the soft thresholding (shrinkage) operator \cite{SToperator}.
After updating $Z,z$ and $\aleph (\dot \alpha )$, the multipliers $m_1,M_2$  can be updated by
\begin{equation} \label{eq_QARsof6}
\begin{array}{l}
{m_1} = {m_1} + u(\aleph (\dot x) - \Re (\dot X)\aleph (\dot \alpha ) - z)\\
{M_2} = {M_2} + u(Z - \Re (\dot X)Diag(\aleph (\dot \alpha )))
\end{array}
\end{equation}
We update the parameter $u$ by $u = \min (\rho u,{u_{\max }})$, where $\rho ,{u_{\max }}$ are two constant values.

\subsection{Classification}
After obtaining the results of the optimization problem of the objective function, we calculate the distance between testing sample $x$ and each class by
\begin{equation}
{d_c} = ||\aleph (\dot x) - \Re ({\dot X_c})\aleph ({\dot \alpha _c})||
\end{equation}
Finally, the testing sample will be classified into the class with the minimization distance by
\begin{equation}
c* = \arg \min ({d_c})
\end{equation}
The detailed classification procedures of QAR are summarized in Algorithm 1.

\begin{figure} [t]
\begin{center}
\subfigure{\includegraphics[width=3.2in]{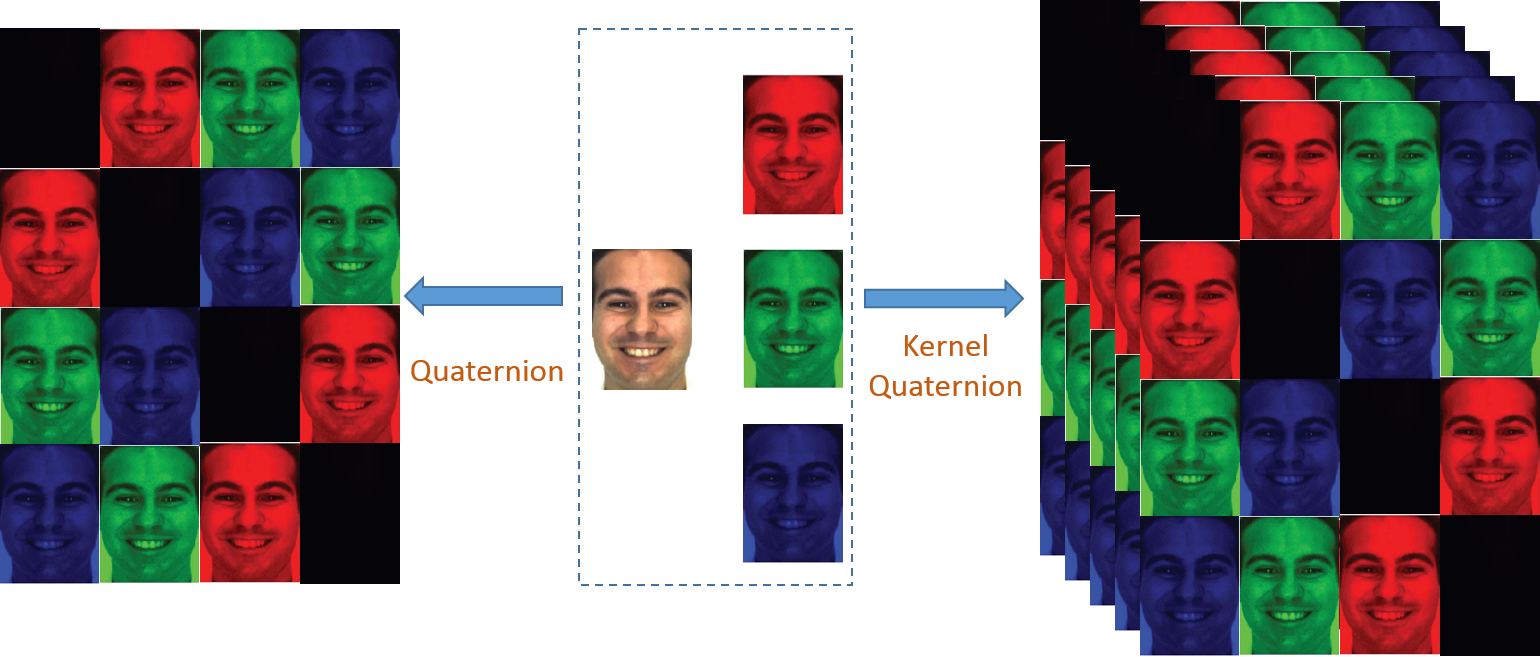}}
\end{center}
\vspace{-0.05in} \caption{ Motivation of HD-AQR. Given a color face image, it has red, green and blue channel. Quaternion transformation can obtain the correlation information among different color channels. The kernel quaternion transformation can obtain the high dimension correlation information among different color channels.} \label{f_motivationHDQAR}
\end{figure}

\section{Proposed HD-QAR}

This section proposes high-dimension quaternion based adaptive representation (HD-QAR). We first introduce the motivation and its objective function. The solution of this objective function and the classification rule are then presented. Finally, we compare the proposed methods with several state-of-the-art methods.

\subsection{Motivation and Objective Function}
To better describe the motivation, we give an example of color face images shown in Fig. \ref{f_motivationHDQAR}. A color face image has red, green and blue channels. The quaternion can obtain structural correlation information among different color channels.  \cite{KSR,KSR1,KSRC} presented high dimension information in kernel space are helpful for sparse. Thus, we want to design a kernel quaternion transformation that can obtain the high dimension correlation information among different color channels.
Suppose that there is a mapping function, $\Phi (.):{{\rm{R}}^q} \to {{\rm{R}}^Q}(q <  < Q)$ , which maps  a vector-based quaternion $\dot x$  and a matrix-based quaternion $\dot X$  to the high dimensional feature space as  $x \to \Phi (\dot x){\kern 1pt} {\kern 1pt} {\kern 1pt} {\kern 1pt} {\kern 1pt} {\kern 1pt} {\kern 1pt} {\kern 1pt} {\kern 1pt} {\kern 1pt} {\kern 1pt} {\kern 1pt} {\kern 1pt} X \to \Phi (\dot X)$ . Using this mapping function, the objective function of HD-QAR is as follows.
\begin{equation} \label{eq_HDQARof1}
\min \;||\dot \alpha |{|_p}\;\;\;\;s.t.\;\;\;\Phi (\dot x) = \Phi (\dot X)\dot \alpha
\end{equation}
where $1 \le p \le 2$.

\subsection{Two kernel-based quaternion operations}
Given a matrix-based quaternion $\dot V = {V_0} + {V_1}i + {V_2}j + {V_3}k$ , the high dimension matrix quaternion is $\phi (\dot V) = \phi ({V_0}) + \phi ({V_1})i + \phi ({V_2})j + \phi ({V_3})k$ . Given a vector-based quaternion $\dot v = {v_0} + {v_1}i + {v_2}j + {v_3}k$ , the high dimension matrix quaternion is $\phi (\dot v) = \phi ({v_0}) + \phi ({v_1})i + \phi ({v_2})j + \phi ({v_3})k$ . Because $\phi (\dot V)$ and $\phi (\dot v)$ contain the complex number, we need to design two operations to transform $\phi (\dot V)$ and $\phi (\dot v)$ to the real number such that the optimization problem in Eq.(\ref{eq_HDQARof1}) can be computed easily. These operations are
\begin{equation} \label{eq_HDQARqo1}
{\Re ^\phi }(\dot V): = \left[ {\begin{array}{*{20}{c}}
{\phi ({V_0})}&{ - \phi ({V_1})}&{ - \phi ({V_2})}&{ - \phi ({V_3})}\\
{\phi ({V_1})}&{\phi ({V_0})}&{ - \phi ({V_3})}&{\phi ({V_2})}\\
{\phi ({V_2})}&{\phi ({V_3})}&{\phi ({V_0})}&{ - \phi ({V_1})}\\
{\phi ({V_3})}&{ - \phi ({V_2})}&{\phi ({V_1})}&{\phi ({V_0})}
\end{array}} \right]
\end{equation}
and
\begin{equation} \label{eq_HDQARqo2}
{\aleph ^\phi }\left( {\dot v} \right) = {\left[ {\begin{array}{*{20}{c}}
{\phi (v_0^T)}&{\phi (v_1^T)}&{\phi (v_2^T)}&{\phi (v_3^T)}
\end{array}} \right]^T}
\end{equation}

\begin{algorithm*}[htbp]
  \begin{algorithmic}[1]
  \REQUIRE The original training set $X$ with $L$ samples and a testing sample  $x$. Initialize the value of $S,s,\alpha ,{m_1},{M_2},u,\rho ,\varepsilon $ and $u_{max}$
  \ENSURE  Label of the testing sample $x$.
  \vspace{3pt}
  \STATE  Use $X$ and $x$ to constitute quaternion-based kernel matrix and kernel vector as
  \[{K_{}} = \left[ {\begin{array}{*{20}{c}}
{k\left( {{\aleph ^\phi }(\dot x_1^{}),{\aleph ^\phi }(\dot x_1^{})} \right)}&{...}&{k\left( {{\aleph ^\phi }(\dot x_1^{}),{\aleph ^\phi }(\dot x_L^{})} \right)}\\
{...}&{...}&{...}\\
{k\left( {{\aleph ^\phi }(\dot x_L^{}),{\aleph ^\phi }(\dot x_1^{})} \right)}&{...}&{k\left( {{\aleph ^\phi }(\dot x_L^{}),{\aleph ^\phi }(\dot x_L^{})} \right)}
\end{array}} \right]~~~~~~~~k = \left[ {\begin{array}{*{20}{c}}
{k\left( {{\aleph ^\phi }(\dot x_1^{}),{\aleph ^\phi }(\dot x)} \right)}&{...}&{k\left( {{\aleph ^\phi }(\dot x_L^{}),{\aleph ^\phi }(\dot x)} \right)}
\end{array}} \right]\]
  \STATE	Use $K$ and $k$ to constitute the objective function as
  \[\mathop {\min }\limits_{S,s,\gamma } ||S|{|_*} + ||s|{|_1}{\kern 1pt} {\kern 1pt} {\kern 1pt} {\kern 1pt} {\kern 1pt} {\kern 1pt} {\kern 1pt} {\kern 1pt} {\kern 1pt} {\kern 1pt} {\kern 1pt} {\kern 1pt} {\kern 1pt} {\kern 1pt} {\kern 1pt} {\kern 1pt} {\kern 1pt} \begin{array}{*{20}{c}}
{s.t.}&{s = k - }
\end{array}K\gamma \begin{array}{*{20}{c}}
{{\kern 1pt} {\kern 1pt} {\kern 1pt} {\kern 1pt} {\kern 1pt} {\kern 1pt} {\kern 1pt} {\kern 1pt} {\kern 1pt} {\kern 1pt} }&{S = KDiag(\gamma )}
\end{array}\]
  \STATE Transform the objective function to a Lagrange multiplier problem as
 \[L(S,s,\gamma ) = \lambda ||S|{|_*} + ||s|{|_1} + y_1^T(k - K\gamma  - s){\kern 1pt} {\kern 1pt}  + Tr[Y_2^T(S - KDiag(\gamma ))]{\kern 1pt} {\kern 1pt}  + \frac{u}{2}(||k - K\gamma  - s||_2^2 + ||S - KDiag(\gamma )||_F^2)\]
  \WHILE {until convergence}
  \vspace{3pt}
  \STATE  Update $S$ by fixing the others with $S* = \arg \mathop {\min }\limits_S \frac{\lambda }{u}||S|{|_*} + \frac{1}{2}||S - (KDiag(\gamma ) - \frac{1}{u}M_2^{})||_F^2$
  \vspace{3pt}
  \STATE  Update $\gamma$  by fixing the others with \[\gamma * = {({K^T}K + Diag(diag({K^T}K)))^{ - 1}}{K^T}(\frac{1}{u}{m_1} + K(k - s)) + {({K^T}K + Diag(diag({K^T}K)))^{ - 1}}diag({K^T}(\frac{1}{u}{M_2} + S))\]
  \STATE  Update $s$ by fixing the others with $s* = \arg \mathop {\min }\limits_s \frac{1}{u}||s|{|_1} + \frac{1}{2}||s - (x - K\gamma  + \frac{1}{u}{m_1})||_2^{}$
  \vspace{3pt}
  \STATE  The multipliers $m_1,M_2$ are updated by ${m_1} = {m_1} + u(k - K\gamma  - s)$ and ${M_2} = {M_2} + u(S - KDiag(\gamma ))$
  \vspace{3pt}
  \STATE   Update the parameter $u$ by $u = \min (\rho u,{u_{\max }})$
  \vspace{3pt}
  \STATE   Check convergence by $||k - K\gamma  - s|{|_\infty } \le \varepsilon $ and $||S - KDiag(\gamma )|{|_\infty } \le \varepsilon $
  \vspace{3pt}
  \ENDWHILE
  \vspace{3pt}
  \STATE   Compute the distance $d_c^{} = \frac{{||k - {K_c}{\alpha _c}||}}{{||{\alpha _c}||}}$ and classify the testing sample $x$  by $c* = \arg \min ({d_c})$

\end{algorithmic}
\caption{Proposed High-Dimension Quaternion-based Adaptive Representation (HD-QAR) Method}\label{algorithm2}
\end{algorithm*}

\subsection{Solution of objective function}

Using the proposed two kernel-based quaternion operations, the proposed objective function of HD-QAR can be represented as.
\begin{equation} \label{eq_HDQARof2}
\begin{array}{l}
\min ||{\Re ^\phi }(\dot X)Diag(\gamma )|{|_*}\\
\begin{array}{*{20}{c}}
{s.t.}&{{\aleph ^\phi }\left( {\dot x} \right) = }
\end{array}{\Re ^\phi }(\dot X)\gamma
\end{array}
\end{equation}
Because the product of high-dimension quaternion are non-communicative, solving the high-dimension quaternion-based optimization problem in HD-QAR is quite challenging.  Thus, we propose quaternion-based kernel matrix $K$ and quaternion-based kernel vector $k$ as follows. Suppose that $K = {\Re ^\phi }{(\dot X)^T}{\Re ^\phi }(\dot X)$ and $k = {\Re ^\phi }{(\dot X)^T}{\aleph ^\phi }(\dot x)$, which can be computed as follows:
\begin{equation} \label{eq_HDQARkm}
\begin{array}{l}
{K_{}} = {\Re ^\phi }{(\dot X)^T}{\Re ^\phi }(\dot X)\\
 = \left[ {\begin{array}{*{20}{c}}
{k\left( {{\aleph ^\phi }(\dot x_1^{}),{\aleph ^\phi }(\dot x_1^{})} \right)}&{...}&{k\left( {{\aleph ^\phi }(\dot x_1^{}),{\aleph ^\phi }(\dot x_L^{})} \right)}\\
{...}&{...}&{...}\\
{k\left( {{\aleph ^\phi }(\dot x_L^{}),{\aleph ^\phi }(\dot x_1^{})} \right)}&{...}&{k\left( {{\aleph ^\phi }(\dot x_L^{}),{\aleph ^\phi }(\dot x_L^{})} \right)}
\end{array}} \right]
\end{array}
\end{equation}
and
\begin{equation} \label{eq_HDQARkv}
\begin{array}{l}
k = {\Re ^\phi }{(\dot X)^T}{\aleph ^\phi }(\dot x)\\
 = \left[ {\begin{array}{*{20}{c}}
{k\left( {{\aleph ^\phi }(\dot x_1^{}),{\aleph ^\phi }(\dot x)} \right)}&{...}&{k\left( {{\aleph ^\phi }(\dot x_L^{}),{\aleph ^\phi }(\dot x)} \right)}
\end{array}} \right]
\end{array}
\end{equation}
Using the Eqs.(\ref{eq_HDQARkm}) and (\ref{eq_HDQARkv}), the Eq.(\ref{eq_HDQARof2}) is rewritten as
\begin{equation} \label{eq_HDQARof3}
\begin{array}{l}
\mathop {\min }\limits_{S,s,\gamma } ||S|{|_*} + ||s|{|_1}\\
\begin{array}{*{20}{c}}
{s.t.}&{s = k - }
\end{array}K\gamma \\
\begin{array}{*{20}{c}}
{{\kern 1pt} {\kern 1pt} {\kern 1pt} {\kern 1pt} {\kern 1pt} {\kern 1pt} {\kern 1pt} {\kern 1pt} {\kern 1pt} {\kern 1pt} }&{S = KDiag(\gamma )}
\end{array}
\end{array}
\end{equation}
The minimization problem in Eq.(\ref{eq_HDQARof3}) is then transformed into the following augmented Lagrange multiplier problem as
\begin{equation} \label{eq_HDQARsof1}
\begin{array}{l}
L(S,s,\gamma ) = \lambda ||S|{|_*} + ||s|{|_1} + y_1^T(k - K\gamma  - s)\\
{\kern 1pt} {\kern 1pt} {\kern 1pt} {\kern 1pt} {\kern 1pt} {\kern 1pt} {\kern 1pt} {\kern 1pt} {\kern 1pt} {\kern 1pt} {\kern 1pt} {\kern 1pt} {\kern 1pt} {\kern 1pt} {\kern 1pt} {\kern 1pt} {\kern 1pt} {\kern 1pt} {\kern 1pt} {\kern 1pt} {\kern 1pt} {\kern 1pt} {\kern 1pt} {\kern 1pt} {\kern 1pt} {\kern 1pt} {\kern 1pt} {\kern 1pt} {\kern 1pt} {\kern 1pt} {\kern 1pt} {\kern 1pt} {\kern 1pt} {\kern 1pt} {\kern 1pt} {\kern 1pt} {\kern 1pt} {\kern 1pt} {\kern 1pt} {\kern 1pt}  + Tr[Y_2^T(S - KDiag(\gamma ))]\\
{\kern 1pt} {\kern 1pt} {\kern 1pt} {\kern 1pt} {\kern 1pt} {\kern 1pt} {\kern 1pt} {\kern 1pt} {\kern 1pt} {\kern 1pt} {\kern 1pt} {\kern 1pt} {\kern 1pt} {\kern 1pt} {\kern 1pt} {\kern 1pt} {\kern 1pt} {\kern 1pt} {\kern 1pt} {\kern 1pt} {\kern 1pt} {\kern 1pt} {\kern 1pt} {\kern 1pt} {\kern 1pt} {\kern 1pt} {\kern 1pt} {\kern 1pt} {\kern 1pt} {\kern 1pt} {\kern 1pt} {\kern 1pt} {\kern 1pt} {\kern 1pt} {\kern 1pt} {\kern 1pt} {\kern 1pt} {\kern 1pt} {\kern 1pt} {\kern 1pt}  + \frac{u}{2}(||k - K\gamma  - s||_2^2 + ||S - KDiag(\gamma )||_F^2)
\end{array}
\end{equation}
where $u>0$ is a parameter, $m_1$  and $M_2$ are Lagrange multipliers. Similar to the proposed QAR, variables $S,s$ and $\gamma$ in Eq.(\ref{eq_HDQARsof1}) can be optimized alternatively when the other two variables are fixed.\\
Update $S$ when $s$ and $\gamma$ are fixed, which is equivalent to solve the minimization problem
\begin{equation} \label{eq_HDQARsof2}
\begin{array}{l}
S* = \arg \mathop {\min }\limits_E L(S,s,\gamma )\\
 = \arg \mathop {\min }\limits_S \lambda ||S|{|_*} + Tr(M_2^TE) + \frac{u}{2}||S - KDiag(\gamma )||_F^2\\
 = \arg \mathop {\min }\limits_S \frac{\lambda }{u}||S|{|_*} + \frac{1}{2}||S - (KDiag(\gamma ) - \frac{1}{u}M_2^{})||_F^2
\end{array}
\end{equation}

\vspace{0.0in}
\begin{figure*} 
\begin{center}
\subfigure{\includegraphics[width=3.2in]{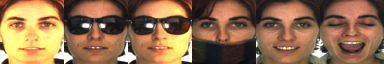}}
\subfigure{\includegraphics[width=3.2in]{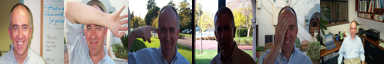}}
\vspace{-0.0in}
\centerline{(a)~~~~~~~~~~~~~~~~~~~~~~~~~~~~~~~~~~~~~~~~~~~~~~~~~~~~~~~~~~~~~~~~~(b)~~~~~~~~}
\subfigure{\includegraphics[width=3.2in]{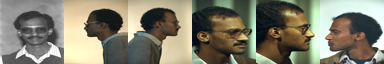}}
\subfigure{\includegraphics[width=3.2in]{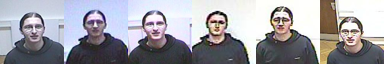}}
\vspace{-0.0in}
\centerline{(c)~~~~~~~~~~~~~~~~~~~~~~~~~~~~~~~~~~~~~~~~~~~~~~~~~~~~~~~~~~~~~~~~~(d)~~~~~~~~}
\subfigure{\includegraphics[width=3.2in]{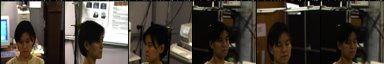}}
\subfigure{\includegraphics[width=3.2in]{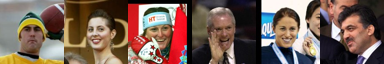}}
\vspace{-0.0in}
\centerline{(e)~~~~~~~~~~~~~~~~~~~~~~~~~~~~~~~~~~~~~~~~~~~~~~~~~~~~~~~~~~~~~~~~~(f)~~~~~~~~}
\end{center}
\vspace{-0.05in} \caption{ Several color images from the (a) AR database, (b) Caltech database, (c) colorFERET database, (d) SCface database, (e) CMU-PIE database and (f) LFW database.} \label{fig_sampleimages}
\end{figure*}
\vspace{0.0in}

\begin{table*} [t]
\caption{The propose Methods vs well-known sparse methods.} \label{t_HDQARvsOther}
\begin{center}
\begin{tabular}{|l|l|l|}
\hline
Methods & Objective Function & Solved minimization problem\\
\hline\hline
SRC  & $\min \;||\alpha |{|_1}\;\;\;\;s.t.\;\;\;x = X\alpha $ &   $e_1$-norm minimization  \\
CRC & $\min \;||\alpha |{|_2}\;\;\;\;s.t.\;\;\;x = X\alpha $ &   $e_2$-norm minimization  \\
KSRC  & $\min \;||\alpha |{|_1}\;\;\;\;s.t.\;\;\;\phi (x) = \phi (X)\alpha $ &   high-dimension-based $e_1$-norm minimization   \\
KCRC  & $\min \;||\alpha |{|_2}\;\;\;\;s.t.\;\;\;\phi (x) = \phi (X)\alpha $  &  high-dimension-based $e_2$-norm minimization   \\
QSRC  & $\min \;||\alpha |{|_1}\;\;\;\;s.t.\;\;\;\dot x = \dot X\alpha $ &   quaternion-based $e_1$-norm minimization  \\
QCRC & $\min \;||\alpha |{|_2}\;\;\;\;s.t.\;\;\;\dot x = \dot X\alpha $  &   quaternion-based $e_2$-norm minimization \\
\hline
QAR & $\min \;||\dot X\alpha |{|_*}\;\;\;\;s.t.\;\;\;\dot x = \dot X\alpha $  &   quaternion-based $e_p$-norm ($1<p<2$) minimization \\
HD-QAR  & $\min \;||\phi (\dot X)\alpha |{|_*}\;\;\;\;s.t.\;\;\;\phi (\dot x) = \phi (\dot X)\alpha $ &  high-dimension-quaternion-based $e_p$-norm ($1<p<2$) minimization  \\
\hline
\end{tabular}
\end{center}
\end{table*}

We use the singular value thresholding (SVT) operator \cite{SVToperator} to approximately solve the minimization problem in Eq.(\ref{eq_HDQARsof2}).\\
Afterwards, update $\gamma$  when $s$ and $S$ are fixed, which can be solved by the following minimization problem
\begin{equation} \label{eq_HDQARsof3}
\begin{array}{l}
\gamma * = \arg \mathop {\min }\limits_\gamma  L(S,s,\gamma )\\
 = \arg \mathop {\min }\limits_\gamma   - m_1^TK\gamma  - Tr(M_2^TKDiag(\gamma ))\\
 + \frac{u}{2}({\gamma ^T}{K^T}K\gamma  - 2{(k - s)^T}K\gamma  - 2Tr({S^T}KDiag(\gamma ))\\
 + Tr({(KDiag(\gamma ))^T}KDiag(\gamma )))\\
 = \arg \mathop {\min }\limits_\gamma  \frac{u}{2}{\gamma ^T}({K^T}K + Diag(diag({K^T}K)))\gamma  - \\
{({K^T}{m_1} + u{K^T}K(k - s) + diag(M_2^TK + u{S^T}K))^T}\gamma
\end{array}
\end{equation}
The above minimization problem can be easily solved by
\begin{equation} \label{eq_HDQARsof4}
\begin{array}{l}
\gamma * = {({K^T}K + Diag(diag({K^T}K)))^{ - 1}}{K^T}(\frac{1}{u}{m_1} + K(k - s))\\
 + {({K^T}K + Diag(diag({K^T}K)))^{ - 1}}diag({K^T}(\frac{1}{u}{M_2} + S))
\end{array}
\end{equation}
Next, update $s$ when $S$ and $\gamma$ are fixed, which can be solved as follows.
\begin{equation} \label{eq_HDQARsof5}
\begin{array}{l}
s* = \arg \mathop {\min }\limits_s L(S,s,\gamma )\\
 = \arg \mathop {\min }\limits_s ||s|{|_1} - m_1^Ts + \frac{u}{2}||k - K\gamma  - s||_F^2\\
 = \arg \mathop {\min }\limits_s \frac{1}{u}||s|{|_1} + \frac{1}{2}||s - (k - K\gamma  + \frac{1}{u}{m_1})||_2^2
\end{array}
\end{equation}
We use the soft thresholding (shrinkage) operator \cite{SToperator} to solve the solution of the minimization problem in Eq.(\ref{eq_HDQARsof5}).
After updating $S$, $s$ and $\gamma$, the multipliers $m_1,M_2$  are updated by
\begin{equation} \label{eq_HDQARsof6}
\begin{array}{l}
{m_1} = {m_1} + u(k - K\gamma  - s)\\
{M_2} = {M_2} + u(S - KDiag(\gamma ))
\end{array}
\end{equation}
Then, update $u$ by  $u = \min (\rho u,{u_{\max }})$, where  $\rho ,{u_{\max }}$ are two constants.

\subsection{Classification}

Using the sparse coefficient $\gamma$, we compute the distance between testing sample and the $c^{th}$ class as
\begin{equation}
d_c^{}(x) = \frac{{||k - {K_c}{\alpha _c}||}}{{||{\alpha _c}||}}
\end{equation}
The rule of HD-QAR is to select the class with the minimum one as
\begin{equation}
\mathop {\min }\limits_{{c^*}} d_c^{}(x)\begin{array}{*{20}{c}}
,&{c = 1,2,...,M}
\end{array}
\end{equation}
The complete classification procedure of HD-QAR is shown in Algorithm 2.

\subsection{Proposed methods vs Well-known methods}
The main difference between the proposed methods and the well-known methods is the way how to obtain the sparse coefficients. Table \ref{t_HDQARvsOther} summarizes and compares the proposed QAR and HD-QAR with these methods.. SRC \cite{SRC} solves $e_1$-norm minimization problem. CRC \cite{CRC} solves $e_2$-norm minimization problem. KSRC \cite{KSR,KSR1,KSRC} solve high-dimension-based $e_1$-norm minimization problem. KCRC \cite{KCRC} solves high-dimension-based $e_2$-norm minimization problem. QSRC \cite{QCRC-QSRC} solves quaternion-based $e_1$-norm minimization problem. QCRC \cite{QCRC-QSRC} solves quaternion-based $e_2$-norm minimization problem. The proposed QAR solves quaternion-based $e_p$-norm ($1<p<2$) minimization problem while the proposed HD-QAR solves high-dimension-quaternion-based $e_p$-norm ($1<p<2$) minimization problem. \cite{QCRC-QSRC} also compares difference between the quaternion-based sparse methods with the group-based sparse representation classification (GSRC) methods \cite{GSR01,GSR02,GSR03}, joint-based sparse representation classification (JSRC) methods \cite{JSR01,JSR02}.

\section{Experiment Results}

This section assesses the effectiveness of the proposed methods. We compare the proposed methods with CRC, ECRC-M, ECRC-A, QCRC, KCRC, SRC, ESRC-M, ESRC-A, GSRC, JSRC, QSRC and KSRC on several well-known color face databases.

\vspace{0.0in}
\begin{figure*} 
\begin{center}
\subfigure{\includegraphics[width=3.2in]{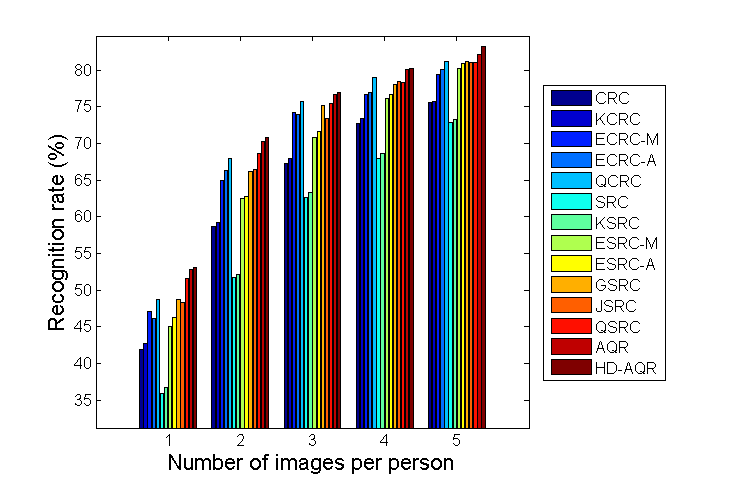}}
\subfigure{\includegraphics[width=3.2in]{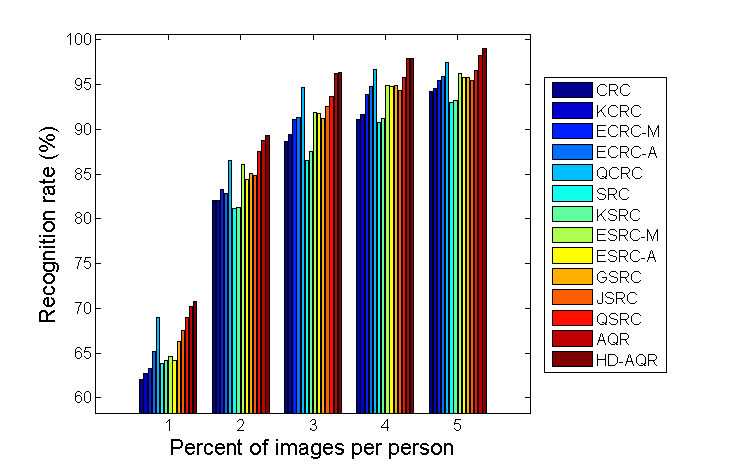}}
\vspace{-0.0in}
\centerline{(a)~~~~~~~~~~~~~~~~~~~~~~~~~~~~~~~~~~~~~~~~~~~~~~~~~~~~~~~~~~~~~~~~~(b)~~~~~~~~}
\subfigure{\includegraphics[width=3.2in]{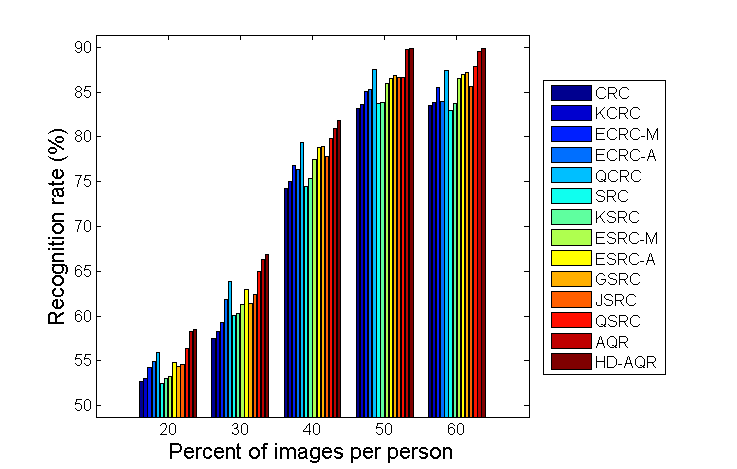}}
\subfigure{\includegraphics[width=3.2in]{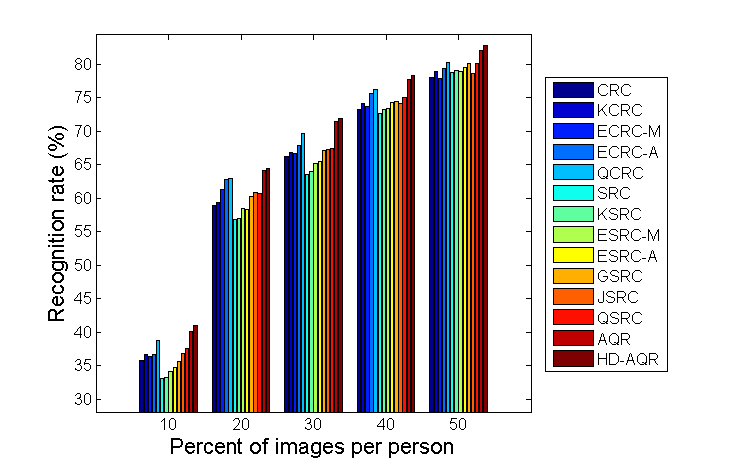}}
\vspace{-0.0in}
\centerline{(c)~~~~~~~~~~~~~~~~~~~~~~~~~~~~~~~~~~~~~~~~~~~~~~~~~~~~~~~~~~~~~~~~~(d)~~~~~~~~}
\end{center}
\vspace{-0.05in} \caption{ The recognition rates of several methods on the (a) AR database, (b) Caltech database, (c) colorFERET database and (d) SCface database.} \label{fig_ARCaltechColorFERETSCface}
\end{figure*}
\vspace{0.0in}

\subsection{Face Recognition with pose, lighting and expressions variations}

AR face database includes 126 people (70 men and 56 women) with more than 4,000 color face images. These face images have different illumination conditions, expressions and occlusions (sun glasses and scarf). Each person has two sessions (2 different days). The same face images were taken in both sessions. The face images were taken at the CVC under strictly controlled conditions. Participants may have different wear (clothes, glasses, etc.), make-up, hair style, etc. Following \cite{QCRC-QSRC}, we select a subset including the images with illumination and expression variations for this experiment. For each person, we randomly choose $n$ (1, 2, 3, 4, 5) face images from session 1 as training set and the face images in session 2 are utilized as testing set. To obtain the convincing results, we randomly select the images and run this process ten times. The average recognition rates of this experiment is described in Fig. \ref{fig_ARCaltechColorFERETSCface} (a), and the mean recognition rates of all $n$ (1, 2, 3, 4, 5) are listed in Table \ref{t_AR}.\\
Caltech Faces Database has 27 subjects with 450 color face images under various illumination, expression and backgrounds. Following \cite{QCRC-QSRC}, we use the Viola-Jones face detector \cite{Viola-Jonesfacedetector} to detect faces. Then, $n$ (1, 2, 3, 4, 5) face images of each person are selected as training set and the rest face images are used as testing set. To get the convincing results, we also randomly select the images and run this process ten times. Fig. \ref{fig_ARCaltechColorFERETSCface} (b) shows the average recognition rates of this experiment. Table \ref{t_Caltech} lists mean recognition rates of all $n$ (1, 2, 3, 4, 5).\\
Color FERET (Facial Recognition Technology) Database has 119 subjects with 14, 126 face images. Under a semi-controlled environment, these images are collected in 15 sessions. Following \cite{QCRC-QSRC}, the frontal face images of the sets with the letter code “fa” and “fb” are utilized for experiments. The Viola-Jones face detector \cite{Viola-Jonesfacedetector} is used to detect faces. In the experiment, we select $p$ (20, 30, 40, 50, 60) percent of face images of each person as training set, and the rest face images are used as testing set. To get the convincing results, we also randomly select the images and run this process ten times. The average recognition rates of this experiment are shown in Fig. \ref{fig_ARCaltechColorFERETSCface} (c). The mean recognition rates of all $p$ (20, 30, 40, 50, 60) are listed in Table \ref{t_colorFERET}.\\
Considering the Tables \ref{t_AR}, \ref{t_Caltech} and \ref{t_colorFERET}, we conclude three points as follows.\\
\begin{itemize}
  \item Due to the fact that the quaternion can obtain the correlation information among different channels of color images, the quaternion-based sparse methods obtain the better performance compared to the non-quaternion-based sparse methods.
  \item The high dimension feature in kernel space can help the sparse-based methods obtain better performance. Thus, the kernel-based sparse methods outperform the corresponding linear-based sparse methods.
  \item The adaptive $e_p$-norm ($1 \le p \le 2$) minimization can benefit from both $e_1$-norm minimization and $e_2$-norm minimization according to the correction of the samples in the prototype set $X$. Therefore, the proposed methods obtain better performance than QCRC and QSRC in \cite{QCRC-QSRC}.
\end{itemize}

\begin{table} [t]
\caption{The recognition rates of several methods on AR face database} \label{t_AR}
\begin{center}
\begin{tabular}{|l|c|}
\hline
Methods & Mean Recognition Rate \\
\hline\hline
CRC  & 63.25  \\
KCRC  & 63.81  \\
ECRC-M & 63.81 \\
ECRC-A  & 68.74  \\
QCRC  & 70.53  \\
SRC  & 58.21 \\
KSRC  & 58.84 \\
ESRC-M  & 66.95 \\
ESRC-A  & 67.65 \\
GSRC  & 69.89 \\
JSRC  & 69.56  \\
QSRC  & 71.03  \\
\hline
QAR  & 72.42  \\
HD-QAR  & 72.87  \\
\hline
\end{tabular}
\end{center}
\end{table}

\begin{table} [t]
\caption{The recognition rates of several methods on Caltech face database} \label{t_Caltech}
\begin{center}
\begin{tabular}{|l|c|}
\hline
Methods & Mean Recognition Rate \\
\hline\hline
CRC  & 83.60  \\
KCRC  & 84.09  \\
ECRC-M & 85.40 \\
ECRC-A  & 86.00  \\
QCRC  & 88.87  \\
SRC  & 83.07 \\
KSRC  & 83.47 \\
ESRC-M  & 86.77 \\
ESRC-A  & 86.20 \\
GSRC  & 86.68 \\
JSRC  & 86.95  \\
QSRC  & 88.52  \\
\hline
QAR  & 90.29  \\
HD-QAR  & 90.70  \\
\hline
\end{tabular}
\end{center}
\end{table}

\begin{table} [t]
\caption{The recognition rates of several methods on colorFERET face database} \label{t_colorFERET}
\begin{center}
\begin{tabular}{|l|c|}
\hline
Methods & Mean Recognition Rate \\
\hline\hline
CRC  & 70.24  \\
KCRC  & 70.76  \\
ECRC-M & 72.17 \\
ECRC-A  & 72.49  \\
QCRC  & 74.82  \\
SRC  & 70.73 \\
KSRC  & 71.27 \\
ESRC-M  & 72.92 \\
ESRC-A  & 74.01 \\
GSRC  & 73.77 \\
JSRC  & 73.43  \\
QSRC  & 75.14  \\
\hline
QAR  & 76.97  \\
HD-QAR  & 77.39  \\
\hline
\end{tabular}
\end{center}
\end{table}

\subsection{Face Recognition with low resolution}

In this section, we use the SCface (Surveillance Cameras Face) to assess the effectiveness of the proposed methods for face recognition with low resolution. The SCface database is a dataset of static face images of human. Face images were collected in uncontrolled indoor environment with 5 video surveillance cameras of various qualities. This dataset includes 130 persons with 4160 face images (in visible and infrared spectrum) from different quality cameras mimic the real-world conditions. The sizes of face images are various. Due to some of the surveillance face images are of quite low resolution and quality, this dataset is quite challenging. Following \cite{QCRC-QSRC}, we randomly choose 2080 face images of 130 persons in this experiment. Then, $p$ (10, 20, 30, 40, 50) percent of face images of each person are used as training set, and the rest face images are used as testing set. To achieve the convincing results, we also randomly select the images and run this process ten times. The average recognition rates of this experiment are plotted in Fig. \ref{fig_ARCaltechColorFERETSCface} (d) and the mean recognition rates of all $p$ (10, 20, 30, 40, 50) are listed in Table \ref{t_SCface}. From this table, we can get the following conclusions. Quaternion-based sparse methods outperform the non-quaternion-based sparse methods. Kernel-based sparse methods have better performance than linear-based sparse methods. The adaptive $e_p$-norm ($1<p<2$) minimization can benefit from both $e_1$-norm minimization and $e_2$-norm minimization according to the correction of the samples in the prototype set $X$. Therefore, the proposed methods obtain the better performance than QCRC and QSRC

\begin{table} [t]
\caption{The recognition rates of several methods on SCface face database} \label{t_SCface}
\begin{center}
\begin{tabular}{|l|c|}
\hline
Methods & Mean Recognition Rate \\
\hline\hline
CRC  & 62.46  \\
KCRC  & 63.16  \\
ECRC-M & 63.18 \\
ECRC-A  & 64.47  \\
QCRC  & 65.53  \\
SRC  & 60.96 \\
KSRC  & 61.32 \\
ESRC-M  & 62.02 \\
ESRC-A  & 62.48 \\
GSRC  & 63.50 \\
JSRC  & 63.53  \\
QSRC  & 64.16  \\
\hline
QAR  & 67.15  \\
HD-QAR  & 67.72  \\
\hline
\end{tabular}
\end{center}
\end{table}

\vspace{0.0in}
\begin{figure*} 
\begin{center}
\subfigure{\includegraphics[width=3.2in]{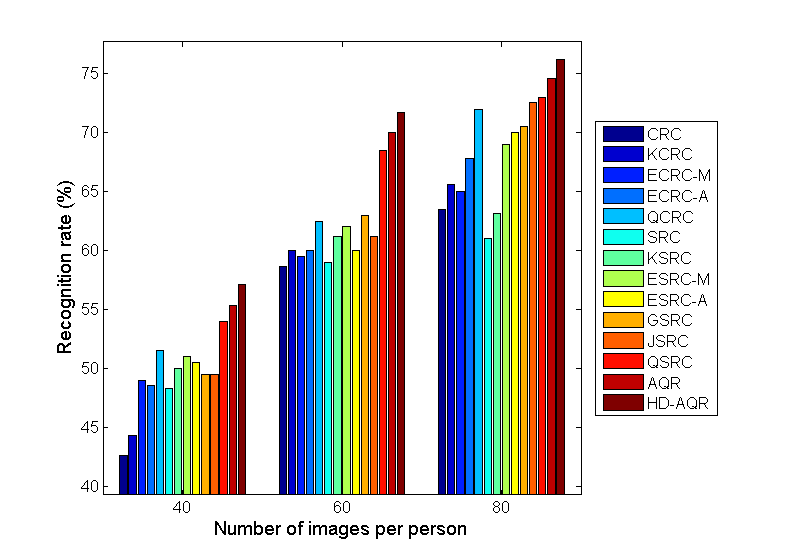}}
\subfigure{\includegraphics[width=3.2in]{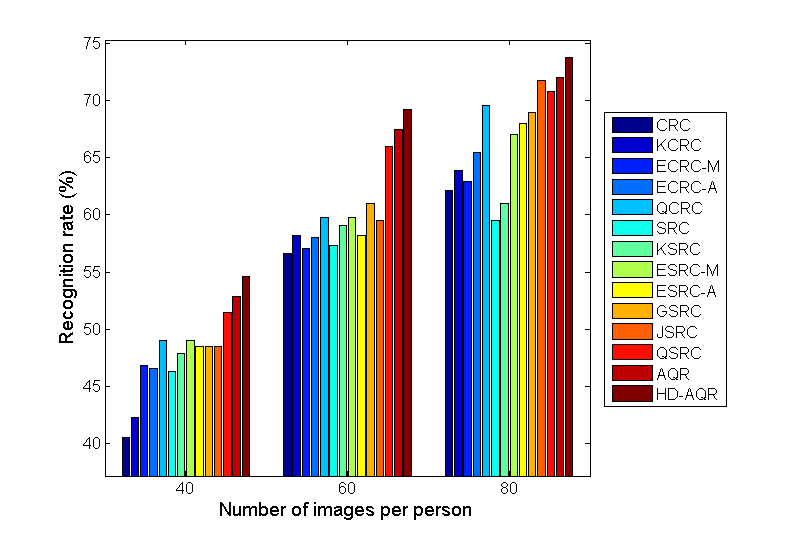}}
\vspace{-0.0in}
\centerline{(a)~~~~~~~~~~~~~~~~~~~~~~~~~~~~~~~~~~~~~~~~~~~~~~~~~~~~~~~~~~~~~~~~~(b)~~~~~~~~}
\end{center}
\vspace{-0.05in} \caption{ The recognition rates of several methods on the (a) LFW face database with normal image and (b) LFW face database with noise image.} \label{fig_LFWnoise}
\end{figure*}
\vspace{0.0in}

\subsection{Face Recognition with complicated background}

This section uses the CMU PIE face database to evaluate the effectiveness of the proposed methods for face recognition with complicated background. CMU PIE database includes more than 40,000 facial images of 68 people. Using the CMU (Carnegie Mellon University) 3D Room, this dataset is collected by imaging each person across 13 different poses, under 43 different illumination conditions, and with four different expressions. This dataset is quite challenging because all face images have the various and complicated background. The experiment is implemented as follows: one-third images of expressions of each person are selected to form training set, while the rest two third images are considered as test images. The experiment results of comparison methods are listed in Table \ref{t_CMUPIE}. We can observe that the proposed methods obtain the better performance than other state-of-the-art methods.

\begin{table} [t]
\caption{The recognition rates of several methods on CMUPIE face database} \label{t_CMUPIE}
\begin{center}
\begin{tabular}{|l|c|}
\hline
Methods & Mean Recognition Rate \\
\hline\hline
CRC  & 75.12  \\
KCRC  & 76.10  \\
ECRC-M & 79.42 \\
ECRC-A  & 78.10  \\
QCRC  & 83.84  \\
SRC  & 77.27 \\
KSRC  & 78.04 \\
ESRC-M  & 84.12 \\
ESRC-A  & 83.79 \\
QSRC  & 85.04 \\
\hline
QAR  & 86.27  \\
HD-QAR  & 86.79  \\
\hline
\end{tabular}
\end{center}
\end{table}

\subsection{Face Recognition in the Wild}

This section uses the LFW (Labeled Faces in the Wild) to assess the performance of the proposed methods for face recognition against environment in the wild. LFW face database has 5749 persons with 13233 color face images taken in real and unconstrained environments. For each person, the face images have large visual variations in illumination, pose, occlusion, etc. This database is quite challenging because the face images of each person have the large variations in unconstrained environments. Following \cite{QCRC-QSRC}, we choose the persons with over $n$ (40,60, 80) face images in this experiment. Then we randomly select 10 percent of the face images per person as training set and the rest images are used as testing set. To get the convincing results, we also randomly select the images and run this process ten times. The average recognition rates of the comparison methods are listed in Fig. \ref{fig_LFWnoise} (a). Observing this figure, we know that the proposed methods could improve the performance of other comparison methods and obtain the higher recognition rate, in spite of the face images having the large variations in unconstrained environments.

\subsection{Face recognition with noise image}

In this subsection, we use the LFW (Labeled Faces in the Wild) face database to test the robustness of the proposed methods against noise. The training sets are same as those of the above section. The testing set is obtained by Matlab function "imnoise" to insert the Gaussian white noise. Notice that only the testing face images are inserted with the Gaussian white noise, and the training face images are utilized the original images of the database. To get the convincing results, we also randomly select the images and run this process ten times. The average recognition rates of the comparison methods are listed in Fig. \ref{fig_LFWnoise} (b). Observing this figure, we know that the proposed methods could outperform other comparison methods and obtain the higher recognition rates for face images with the noise and large variations in unconstrained environments.

\section{Conclusion}
In this paper, quaternion-based adaptive representation (QAR) was proposed for color face recognition. QAR can adaptively combine the $e_1$-norm and $e_2$-norm and benefit from both. Moreover, we proposed the high-dimension quaternion-based adaptive representation (HD-QAR), which can obtain the high dimension feature information in the kernel space, Thus, HD-QAR can obtain the correlation information among the different channels. Experimental results showed that the proposed methods achieve better performance compared to several state-of-the-art methods. The analyses and the experimental results on three databases confirmed the effectiveness of the proposed methods for color face recognition.

{\small
\bibliographystyle{IEEEtran}
\bibliography{egbib}
}

\end{document}